# Stock Market Price Prediction using Neural Prophet with Deep Neural Network

1st Navin Chhibber
*Infinity Tech Group*
Sunnyvale, USA
naveenchibber.research@gmail.com

2nd Sunil Khemka
*Persistent Systems*
Chicago, USA
sunilkhemka.tech@gmail.com

3rd Navneet Kumar Tyagi
*Finance of America*
Rosenberg, USA
navneettyagi.research@gmail.com

4th Rohit Tewari
*Unysis*
Fairfax, USA
rohittewari.fintech@gmail.com

5th Bireswar Banerjee
*VISA*
Austin, USA
bireswar.infosys@gmail.com

6th Piyush Ranjan
*IEEE Vice Chair AeroSpace Chapter*
Edison, USA
piyush.ranjan@ieee.org

*Abstract*—Stock market price prediction is a significant interdisciplinary research domain that depends at the intersection of finance, statistics, and economics. Forecasting stock prices accurately has always been a focal point for various researchers. However, existing statistical approaches for time-series prediction often fail to effectively forecast the probability range of future stock prices. Hence, to solve this problem, the Neural Prophet with a Deep Neural Network (NP-DNN) is proposed to predict stock market prices. The preprocessing technique used in this research is Z-score normalization, which normalizes stock price data by removing scale differences, making patterns easier to detect. Missing value imputation fills gaps in historical data, enhancing the model's use of complete information for more accurate predictions. The Multi-Layer Perceptron (MLP) learns complex nonlinear relationships among stock market prices and extracts hidden patterns from the input data, thereby creating meaningful feature representations for better prediction accuracy. The proposed NP-DNN model attained an accuracy of 99.21% when compared with other existing approaches using the Fused Large Language Model (LLM), respectively.

## I. INTRODUCTION

Stock price prediction is an essential topic in economics because of random stock market volatility. It is often one of the most difficult subjects in stock process forecasting [1], [2]. When investing in any company's stock, there is no alternative way to determine the company's performance; therefore, every individual performs data analysis before investing [3]. The financial stock market produces a huge amount of stock price data daily. Shareholders and scholars have focused on mastering and learning stock price prediction rules using available data [4], [5]. The foundation of financial development and decision-making to improve traders' and investors' interests represents effective financial predictions [6]. A huge amount of stock price data is required for the right investment selection within a short time. Investors and traders' stock price data are significant contributors to instability in the stock market, which leads to major investments and losses [7], [8].

The frequent fluctuation in price rarely remains constant, and the company's success is influenced by its perceived reputation and past performance. Based on demand and supply, stock prices fluctuate —demand increases prices, and low demand reduces them [9], [10]. Stocks are sold and bought on the stock exchange, where buyers and sellers conduct transactions and communicate with each other [11].

Existing approaches, such as exponential smoothing moving changes, and other methods for better data handling, have struggled with complex market dynamics and nonlinear feature capturing in the stock market [12]. Accurate stock price estimation helps formulate strategic decisions for investment planning, enabling organizations to manage business undertakings, mergers, and acquisitions effectively [13]. Various traditional methods lack stock price prediction capability. Artificial Intelligence (AI) with advanced Machine Learning (ML) and Deep Learning (DL) can capture nonadditive and nonlinear relationships in complex financial market data to achieve superior prediction results [14], [15]. The contributions of this study are as follows:

- The Neural Prophet (NP) approach captures historical patterns, trends, and seasonality in stock market data and effectively handles irregular intervals and missing timestamps, providing interpretable forecasts that improve the quality of DNN inputs.

- Deep Neural Networks (DNN) learn complex nonlinear relationships among market features, providing accurate price predictions using extracted features that generalize well to unseen data when properly tuned.

- Multi-Layer Perceptron (MLP) is used for feature extraction, transforming raw tabular data into meaningful representations and capturing hidden patterns that enhance prediction performance, reduce dimensionality and noise, and provide better input to the DNN.

The remainder of this research is organized as follows: Section 2 presents the Literature survey, Section 3 describes the workflow of the proposed method. Section 4 provides the results and discussion, and Section 5 concludes the research.

## II. LITERATURE SURVEY

In this section, the literature review is defined and determined based on the existing research, along with its advantages and limitations.

Zhen Qiu et al. [16] introduced a Decision Support System (DSS) to predict modelling by utilizing supervised learning and considering the historical data of startups that succeeded or failed to secure funding. The DSS was used to visualize predictions and key sentiment drivers to assist startup founders and investors. Startup strategy funders monitored sentiment trends to adjust communication and helped investors identify promising startups early based on sentiment.

However, behaviors such as slang, sarcasm, and cultural nuances distorted the accuracy of sentiment analysis.

Yiea-Funk Te et al. [17] implemented a Light Gradient Boosting Machine (LightGBM) for classification. These models were used to train three models, namely Crunchbase, LinkedIn, and a combined model, and Shapley value analysis was utilized to interpret feature importance. The LightGBM model inherited biases from historical funding patterns, which helped improve the overall system performance. However, this scarcity led to biased model training that reduced predictive accuracy, making it difficult to generalize findings across different sectors or regions.

Arman Arzani et al. [18] developed a Random Forest (RF) to classify individuals as potential founders and used historical data to predict future founding. The RF model assisted innovation coaches in identifying high-potential individuals by prioritizing candidates for entrepreneurial training, mentorship, and incubation programs. An RF approach was trained at the university to generalize well to others and to enable proactive support for budding entrepreneurs. However, the RF model faced the risk of tailoring models too closely to a particular institutional context and ignored soft traits unless the model was explicitly modelled.

Abdurahman Maarouf et al. [19] introduced a Large Language Model (LLM) that combined structured and unstructured data. The structured data contained quantitative features such as number of founders, startups, business factors, and startup age. The unstructured data contained textual self-descriptions from startup profiles on platforms such as Crunchbase. The textual signal was utilized to demonstrate how a startup described itself as a predictor of success. However, startup success factors evolved over time, which required regular model updates, and the data did not perform equally well across other regions or platforms.

Pablo Mac Clay et al. [20] implemented Natural Language Processing (NLP) to analyze and classify company descriptions. The analysis of investment patterns of major agri-food multinationals was classified as defensive, upgrading, and portfolio. NLP provided a clear approach for incumbent firms to understanding their motivations and tactics. However, inferring a corporate firm's investment strategy from the investment patterns was speculative without internal firm data.

### III. PROPOSED METHODOLOGY

In this research, the proposed Neural Prophet with a Deep Neural Network (NP-DNN) is used to predict stock prices based on extracted features and hyperparameter tuning using Optuna, which optimizes DNN parameters to enhance prediction performance. A Crunchbase dataset is used as input for model consistency, and preprocessing techniques such as Z-score normalization are applied to standardize numerical features. Missing value imputation fills incomplete data to ensure model consistency. MLP learns hidden patterns and generates meaningful feature representations. Fig. 1 represents the stock market price prediction process.

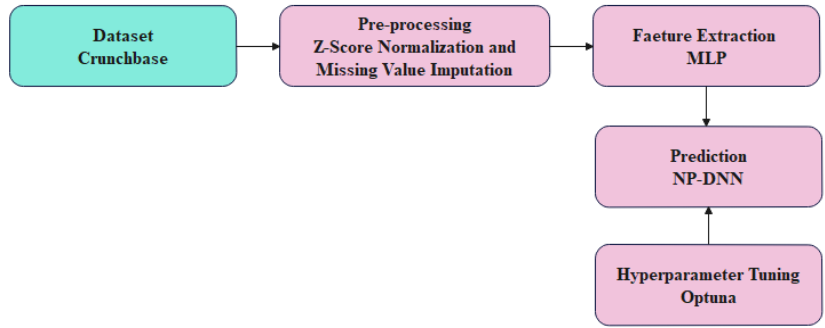

Fig. 1. Workflow of Stock Market Price Prediction

#### A. Dataset Description

The Crunchbase dataset [21] is a publicly available database containing accessible LinkedIn profile data. Crunchbase is a platform for public and private companies that provides business information, professional profiles and founders in leadership, investors, positions, and financing rounds. Experiments and studies of the Crunchbase dataset were conducted in September 2021. This dataset consists of various tables broadly divided into three categories of information: (1) organization, (2) people, and (3) investments. Additionally, the Crunchbase data include sampled LinkedIn profile data for a subset of companies with individuals who provided their LinkedIn profiles on Crunchbase.

#### B. Pre-processing

Crunchbase data are provided as input for preprocessing to convert raw numerical data into formatted data that improve data quality. Early preprocessing of data enhances the accuracy and value of financial information for future analysis. In this research, the problem addresses the deficiency in values while preserving the validity and accuracy of the information. The Z-score technique is also applied by adding components on the same scale to standardize the data, making it easier to identify efficient developments and patterns in financial statistics. Activities such as stock market forecasting, nowcasting, evaluation of decision-making, and risk regarding investments benefit from these preprocessing procedures because of their importance in enhancing the reliability and accuracy of analysis. To replace the observed irregularities by adjusting them to resemble the original data, a two-stage approach is utilized. The panda collection technique is used to apply time-based imputation after eliminating any outliers. The value gap between two data points is determined by linear interpolation. The missing value is presented in Eq. (1).

$$Y = y_1' + (x - x_x')\frac{y_2' - y_1'}{x_2' - x_1'}, \qquad (1)$$

Where $x_1', x_2'$ are the two boundary point time indices are associated with endpoint values and their equivalents $y_1'$ and $y_2'$. The distance between $x_1'$ and $x_2'$ represents the distance. When the imputed data are produced using this method, they deviate less from the average variation of the nearby normal data.

The Z-score normalization approach performs preprocessing for the initial data before analysis by the model. Initially, the data are analyzed for their standard deviation, and Z-score normalization creates a normalized result. The Z-score parameter for raw information normalization is given in Eq. (2).

$$K_n'' = \frac{K_n'' - \bar{F}}{sd}, \quad (2)$$

Where the standard Z-score is denoted as $K_n''$, which represents the normalized value. Here, $K_n''$ and $k_n$ indicates row $F$ of the first column where the value occurs. Since every row outcome is the same, it typically produces data with an average deviation of zero.

*C. Feature Extraction using Multilayer Perceptron*

The pre-processed data are provided as input for feature extraction to improve the prediction accuracy. In an MLP, the minimum number of layers includes an input layer, hidden layer, and output layer, which allows the model to combine learning from incoming data due to nonlinear activation functions and several layers. Essentially, the interdependencies among variables such as trading volume, economic factors, and price stories may be nonlinear. An MLP is capable of modelling these dependencies and nonlinear patterns and extracting features that may be subtle for the analysis. The highly flexible MLP can be fine-tuned for optimal feature extraction based on the unique characteristics of stock market data. Additional hidden layers frequently handle complex approximation tasks in the MLP architecture. A single MLP network with a directional connection composed of several node layers interconnected with subsequent layers that converts an array of input variables into an array of output variables. The MLP is frequently trained using a supervised learning technique called the backpropagation algorithm. The disadvantage of a perceptron, which is its inability to identify linearly indivisible data, is addressed by this extension. The MLP is fully connected with neurons, and each layer is connected to every cell of other layers, representing the total and extra weights.

In particular, $X$ represents the input layer, and $K1$ represents the hidden layer, as given in Eq. (3).

$$K_1 = f(V_1 X + C_1). \quad (3)$$

Where the non-linear activation operation is denoted as $f$, the weight variables are represented as $V_1$, and the bias variables as $C_1$ To obtain the hidden layer output, comparable procedures are performed throughout the hidden layers. Each layer possesses bias settings, and the unique weight from the hidden layer to the output layer is given by Eq. (4).

$$Y = G(V_L K_L + C_L) \quad (4)$$

Where $L$ denotes the hidden layer integer and $K\_L$ represents the final hidden layer. The activation function process, sometimes referred to as SoftMax, is represented as $G$. As a result, the connection between the input and output is developed, and the output of a neuron varies when its bias (c) or weight (v) is slightly altered.

*D. Prediction using DNN*

In this research, a DNN is an activation layer, and dense layers are commonly used to transform the input training data into a high-dimensional feature space. Effective learning helps capture the complex features of the underlying stock market price patterns from data. A linear transformation is applied from the dense layer to the input data by multiplying it by the added weight and bias, as given in Eq. (5).

$$y = x_n . w_n + b_n \quad (5)$$

The SoftMax layer is used for the output layer, which is a popular logistic function that accepts values between 0 and 1. The value closest to 1 represents the final outcome and the value of the neurons derived from the $z$ output layer, as given in Eq. (6).

$$Softmax(z_i) = \frac{e^{zj}}{\sum_j e^{zj}} \quad (6)$$

To prepare data for the dense layer that follows from the flattened layer, the flattened layers manage this conversion by flattening the output, yielding a one-dimensional representation.

*1) Neural Prophet (NP):* This NP model is combined with DL to improve prediction and has multiple elements such as trend, seasonality, extra regression, and auto-regression. Three essential components of the holiday prophet are trends and seasonality. The aforementioned components are integrated into Eq. (7).

$$NP(t) = TMF(t) + SF(t) + HF(t) + EV(t) \quad (7)$$

Where $SF(t)$ denotes seasonality, $HF(t)$ is the holiday function, and $EV(t)$ represents the error variation.

*2) Hyperparameter Tuning using Optuna:* The Optuna capabilities, a Bayesian optimization strategy used for hyperparameter tuning, help in the manual selection of intelligent, hyperparameter, and systematic optimization processes. Optuna performs an intricate search process for hyperparameter exploration within predefined ranges. A significant boost in the performance of the proposed DNN model is achieved by this automated tuning mechanism. The Optuna methodically transverses the hyperparameter landscape, ensuring that the DNN model is optimally configured to precisely execute stock price predictions.

IV. EXPERIMENT RESULTS

The performance of the NP-DNN is determined based on the dataset used for training and testing. The developed NP-DNN is implemented in a Python 3.8 environment, and the system runs with an Intel Core i7-4200U CPU processor, Windows 10 (64-bit OS), and 16GB RAM. The developed NP-DNN technique is estimated using various analytical metrics: recall, precision, F1-Score, and the mathematical expressions for each metric are formulated in Eqs. (8)–(11), respectively.

The percentage of correctly predicted values in the dataset is called accuracy.

$$Accuracy = \frac{TP}{TP+FN+TN+FP} \quad (8)$$

The percentage of true positive values among all predicted positive values is called precision.

$$Precision = \frac{TP}{TP+FP} \quad (9)$$

The fraction calculation of all positive samples estimated to be positive using the predictive utility equation is called recall.

$$Recall = \frac{TP}{TP+FN} \quad (10)$$

The F1-Score measures the reciprocity between precision and recall, as shown below.

$$F1 = \frac{2 \times Precision \times Recall}{Precision \times Recall} \quad (11)$$

Where $TP$ – True Positive; $TN$ – True Negative; $FP$ – False Positive; and $FN$ – False Negative.

*A. Performance Analysis*

A comprehensive analysis of the developed NP-DNN is performed using the Crunchbase dataset. The efficiency of the designed approach is validated in this section using various performance metrics. Fig. 2 shows the results of hyperparameter tuning.

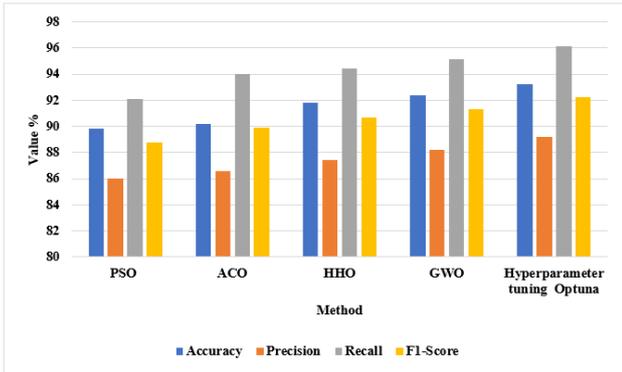

Fig. 2. Analysis of Optuna with existing optimization algorithms

In Fig. 2, the proposed DNN method is represented by various performance metrics using the Crunchbase dataset. The efficiency of different optimization methods, such as Particle Swarm Optimization (PSO), Ant Colony Optimization (ACO), Harris Hawks Optimization (HHO), and Grey Wolf Optimization (GWO), is evaluated to identify the most effective approach for improving model accuracy and convergence. In comparison with these methods, the developed NP-DNN achieves the best exploration capability and completely addresses the features of emotion recognition, thus leading to a significant selection of essential features. The proposed NP-DNN method attained an accuracy of 93.21% for the Crunchbase dataset.

*B. Comparative Analysis*

The proposed NP-DNN method is compared using multiple performance metrics. Fig. 3 shows the comparative outcomes of the developed approach with existing methods, namely DSS [16], LightGBM [17], RF [18], and LLM [19], based on the Crunchbase dataset. Compared to existing models, NP-DNN achieves higher accuracy in predicting stock prices. The method NP-DNN performance is evaluated by utilizing performance metrics.

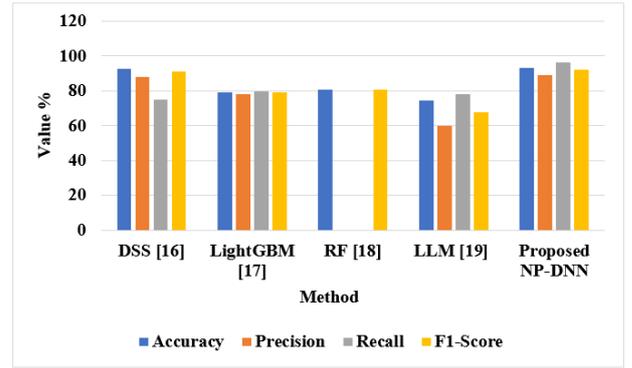

Fig. 3. Comparative Analysis of the Proposed method for RMSE

In Fig. 3, the proposed NP-DNN methodology is introduced along with different performance metrics on the Crunchbase dataset. The Proposed method performs efficiently compared to the analyzed methods, enhancing classification accuracy. The proposed NP-DNN method attained an accuracy of 93.21% for the Crunchbase dataset.

*C. Discussion*

The stock market's inherent volatility, driven by both internal and external factors such as political events, economic shifts, and environmental changes, makes accurate price prediction challenging when relying solely on historical data. Integrating real-time information from microblogs and news sources can provide valuable insights; however, advanced models are required to process and interpret it effectively. The NP-DNN model enhances prediction by capturing dependencies in both past and future data, which is crucial for time-series forecasting. Coupling this with Optuna for hyperparameter tuning significantly improves prediction accuracy by optimizing the model parameters. Additionally, preprocessing with Z-score normalization and missing value imputation helps normalize the values, while the MLP feature extraction method extracts and fills the most relevant data features, leading to more precise predictions. Together, these techniques form a robust framework to improve the accuracy and reliability of stock price predictions in a volatile market. The proposed NP-DNN method attained an accuracy of 93.21% for the Crunchbase dataset.

V. CONCLUSION

Predicting stock market prices accurately is a complex task due to market volatility and the influence of numerous external factors. The combination of optimizing the NP algorithm with NP-DNN for hyperparameter tuning efficiently improves model prediction performance. Z-score normalization helps to scale the values by enhancing data quality, and the MLP feature extraction technique refines the input data and extracts the most important features, enhancing the model's ability to make accurate predictions. The proposed method highlights the importance of optimization algorithms for predicting in unpredictable environments, which is challenging for stock price forecasting. It is difficult to identify stock prices in unpredictable and dynamic environments. Future work will explore the integration of additional financial and macroeconomic variables to enhance model accuracy. Investigating other deep learning architectures or hybrid optimization techniques may further improve the predictive performance.